\documentclass[runningheads]{llncs}
\usepackage[T1]{fontenc}
\usepackage{graphicx}
\usepackage{booktabs}
\begin{document}
\title{Emotional Support with LLM-based Empathetic Dialogue Generation}
%
%
\author{Shiquan Wang\textsuperscript{*} \and
Ruiyu Fang\textsuperscript{*} \and Zhongjiang He\and Shuangyong Song\and Yongxiang Li\textsuperscript{\dag}}
\authorrunning{S. Wang et al.}

%
%
\institute{Institute of Artificial Intelligence (TeleAI), China Telecom Corp Ltd
\email{\{wangsq23,fangry,hezj,songshy,liyx25\}@chinatelecom.cn}}
%
\maketitle              
\begingroup
\renewcommand\thefootnote{\textsuperscript{*}}
\footnotetext{Equal contribution.}
\renewcommand\thefootnote{\textsuperscript{\dag}}
\footnotetext{Corresponding author.}
\endgroup
\begin{abstract}
Emotional Support Conversation (ESC) aims to provide empathetic and effective emotional assistance through dialogue, addressing the growing demand for mental health support. This paper presents our solution for the NLPCC 2025 Task 8 ESC evaluation, where we leverage large-scale language models enhanced by prompt engineering and fine-tuning techniques. We explore both parameter-efficient Low-Rank Adaptation and full-parameter fine-tuning strategies to improve the model’s ability to generate supportive and contextually appropriate responses. Our best model ranked second in the competition, highlighting the potential of combining LLMs with effective adaptation methods for ESC tasks. Future work will focus on further enhancing emotional understanding and response personalization to build more practical and reliable emotional support systems.

\end{abstract}
\section{Introduction}
Emotional Support Conversation (ESC) aims to alleviate individuals' emotional distress and provide effective emotional support through conversational interactions\cite{song2015detecting,song2019concept,song2021emotion}. With the increasing prevalence of mental health issues globally, ESC has demonstrated significant potential as an intervention tool in various domains, including mental health support, customer service, and interpersonal interactions\cite{jiang2025utterance,jiang2025label,kang2024can,liu2021towards,shi2025muddit,liu2023icsberts,song2021emotional,song2020sentiment,wang2024towards,ouyang2023mining,xiong2024dual,yao2020session,song2015classifying,wu2025table,zhang2024lemur,song2020tcnn,song2020two,wu2025mr,li2024scalable,ning2023ump,zhao2023muse,song2015recommending}. ESC systems not only need to process information from everyday conversations but must also focus on understanding and addressing users' emotional needs, providing targeted emotional responses. As such, the design of emotional support dialogue systems faces several challenges, such as accurately understanding subtle emotional fluctuations in users, generating emotionally authentic and constructive responses, and designing support strategies that are adaptable to different contexts and user needs\cite{zhou2023facilitating}.

Despite significant advancements in sentiment analysis and emotional response generation technologies in recent years, existing emotional support dialogue systems still have considerable limitations. Many systems struggle to effectively capture the nuances and complexity of emotions, and the generated responses often lack true emotional resonance and practical help\cite{li2024helpful,shao2025ai}. This represents a critical area for continued exploration and optimization in the ESC field. To further advance research in this domain, the NLPCC 2025 Task 8 has organized an Emotional Support Conversation evaluation task, aimed at assessing different models' capabilities to understand and respond to users' emotional needs, providing a unified dataset and evaluation standards.

In this evaluation task, our team adopted a fine-tuning approach based on large-scale language models (LLMs) and achieved second place \cite{liu2024icsplms,wang2024telechat}. LLMs were selected as the core models due to their remarkable capabilities in natural language understanding and generation, making them ideal for the ESC task. By fine-tuning these models on a specialized emotional support dialogue dataset, we aimed to better align the models with the specific demands of ESC tasks, enabling them to generate empathetic and contextually appropriate responses that cater to users' emotional needs.
Specifically, we employed the qwen2.5-72B-instruct and qwen2.5-7B-Instruct models, applying both low-rank adaptation (LoRA) fine-tuning and full-parameter fine-tuning strategies to further improve performance. Our experimental results demonstrate that appropriate fine-tuning techniques can significantly enhance the performance of models in emotional support conversation tasks.

\section{Related Work}
Emotional Support Conversation (ESC) has become an important research direction within the field of Natural Language Processing (NLP), attracting significant attention from researchers. Existing studies can be broadly categorized into the following approaches based on the techniques employed:

\subsection{Rule-Based Approaches}
Rule-based ESC systems typically rely on manually designed rules and templates to generate supportive responses. These systems identify keywords or patterns in the user's input and select appropriate responses from a predefined list. For example, a system may recognize a specific topic mentioned by the user and respond using a predefined template or scripted message~\cite{chu2024towards}. Early conversational agents, such as ELIZA, were built using rule-based methods, providing simple emotional support through scripted dialogue. However, rule-based approaches have limitations in capturing the subtle nuances of human emotions, resulting in responses that may feel rigid and predictable~\cite{alazraki2021empathetic,yu2024experimental}. While these methods offer certain advantages in terms of safety and controllability—ensuring that inappropriate responses are avoided—they lack the flexibility and depth required to provide truly empathetic support. The inability to deeply understand the user's emotional state remains a major constraint.

\subsection{Retrieval-Based Approaches}
Retrieval-based ESC systems generate responses by selecting the most relevant reply from a predefined corpus. These methods often rely on keyword matching or machine learning techniques to find the most similar response to the user's input. Retrieval-based systems have advantages in ensuring the coherence and computational efficiency of responses~\cite{jia2023knowledge,song2017intention,wang2022topicks,xu2024dynamic}. Some studies have also explored incorporating techniques such as emotional clustering and explainable AI into retrieval-based ESC systems to enhance their effectiveness. However, the performance of retrieval-based methods is heavily dependent on the quality and coverage of the predefined corpus~\cite{zhao2023transesc,zheng2024self}. As a result, they may struggle to provide appropriate responses to novel situations or scenarios that were not anticipated in the corpus.

\subsection{Generation-Based Approaches}
Generation-based ESC systems, particularly those employing neural networks and large-scale language models (LLMs), are capable of generating entirely new responses. These approaches train models on vast amounts of conversational data, enabling them to understand context and generate natural language responses~\cite{chen2024cauesc,cheng2023pal,li2024tele,zhao2024autograph}. However, the use of generative methods in ESC tasks introduces several challenges, such as maintaining conversational coherence and ensuring that the generated emotional responses are appropriate~\cite{kluwer2011chatbots,vanel2023survey}. With the rapid development of LLMs in recent years, there has been increasing exploration into leveraging LLMs and fine-tuning techniques to build more effective ESC systems. LLMs, with their strong generalization ability and capacity to learn complex patterns from large datasets, hold significant promise for enhancing emotional support dialogue systems.

In addition to the aforementioned categories, many studies have focused on how to integrate emotional intelligence and empathy into dialogue systems. This includes research on modeling the user's emotional state and generating responses that demonstrate empathy~\cite{he2024telechat,zhao2025fisminess}. Some studies have emphasized the importance of considering the help-seeker's personal traits when providing effective support. Furthermore, generating inappropriate responses remains a critical issue, and researchers are actively exploring various techniques to mitigate this problem. Building genuinely empathetic ESC systems requires not only understanding the user's emotions but also accounting for their specific circumstances, personality, and the potential for inappropriate responses.

Our work falls under the generation-based approach, focusing on fine-tuning large-scale language models to enhance the capability of ESC systems. In the following sections, we will detail the specific models and fine-tuning strategies we employed, as well as the results achieved in the NLPCC 2025 Task 8 evaluation task.

\section{Methodology}
This section provides a detailed description of the technical approach we adopted for the NLPCC 2025 Task 8 Emotional Support Conversation evaluation. Our overall methodology is based on the capabilities of large language models (LLMs). By combining carefully designed prompt engineering with parameter-efficient fine-tuning strategies, we enable the model to better adapt to emotional support scenarios and generate responses that are both empathetic and practically helpful.

\subsection{Prompt Engineering}

To effectively guide the LLM in understanding user emotions and generating emotionally resonant responses, we designed a structured and semantically clear prompt template. This prompt explicitly defines the model's role, task objective, and response guidelines to ensure that the outputs exhibit key characteristics such as human-likeness, empathy, and personalization. Specifically, the prompt consists of the following components: Role Definition, where the model is positioned as an empathetic assistant responsible for providing sincere, warm, and personalized support based on the user's background and emotional state; Task Objective, which instructs the model to generate supportive responses aimed at relieving negative emotions and boosting emotional resilience, emphasizing relevance to the user's specific context; User Profile, which includes a description of the user's life experience, personality traits, and current emotional struggles, enabling personalized generation; and Response Guidelines, which specify language tone, empathy strategies, and risk avoidance principles to ensure that the generated content remains warm, non-judgmental, and attentive. Through this multi-faceted prompt design, we significantly enhanced the model’s ability to understand the user’s emotional needs and maintain high-quality response generation with strong contextual awareness and emotional control.

\subsection{Model Fine-Tuning with LoRA}
For the core model, we selected Qwen2.5-72B-Instruct and adopted Low-Rank Adaptation (LoRA) as a parameter-efficient fine-tuning strategy to reduce training cost while improving task adaptability. The core idea of LoRA is to insert trainable low-rank matrices into each layer of the model and update only these new parameters, while keeping the original weights frozen. This enables efficient adaptation without modifying the full model. We explored multiple LoRA parameter configurations (e.g., rank=8/16/32 and alpha=16/32/64), and conducted comparative experiments on validation and test sets to identify the optimal setup. Our results demonstrate that appropriate LoRA configurations can significantly enhance the emotional support capabilities of the model while maintaining high efficiency, making it a practical and scalable solution for large-scale emotional dialogue modeling.

\subsection{Full Parameter Fine-Tuning}
In addition to LoRA, we applied full-parameter fine-tuning to the smaller Qwen2.5-7B-Instruct model, where all parameters of the pre-trained model are updated during the training process. Although this approach is more resource-intensive compared to parameter-efficient methods, it allows for greater expressive capacity and better performance, especially when the training data is sufficient. In our study, full fine-tuning served both as a complement and as a comparative baseline to LoRA, enabling us to systematically analyze the advantages and limitations of different fine-tuning strategies in the emotional support task. The experimental results show that full fine-tuning can further improve the model’s ability to recognize and respond to nuanced emotional expressions, thereby producing more delicate and natural emotional support responses.

\section{Experiments}

This section presents the experimental setup, evaluation metrics, and results for the NLPCC 2025 Task 8 on Emotional Support Conversations. Our experiments are designed to systematically verify the effectiveness of our proposed methods in generating emotionally supportive responses.

\subsection{Dataset}

We use the official dataset provided by the NLPCC 2025 Task 8 organizers. The dataset simulates real-world emotional support scenarios, where users (seekers) express emotional distress and the system (supporter) responds with supportive dialogue. Each instance contains multi-turn interactions between the user and the assistant.
The dataset comprises 1,500 dialogue instances, which are split into training, validation, and test sets with an 8:1:1 ratio — resulting in 1,200 training dialogues, 150 validation dialogues, and 150 test dialogues.

\subsection{Evaluation Metrics}

To evaluate the quality and emotional support effectiveness of generated responses, we adopt a range of automatic evaluation metrics:

\begin{itemize}
  \item \textbf{BLEU-4 (B-4)}: Measures n-gram overlap between the generated response and reference;
  \item \textbf{METEOR (ME)}: Accounts for synonymy, stemming, and word order for semantic similarity;
  \item \textbf{ROUGE-L (R-L)}: Based on the longest common subsequence to assess content overlap;
  \item \textbf{Vector Extrema (Ext)}: Uses the extrema of word embeddings to compute semantic similarity;
  \item \textbf{Distinct-2/Distinct-3 (D-2/D-3)}: Evaluate lexical diversity through the ratio of unique bigrams and trigrams;
  \item \textbf{G-Score}: An overall human-centric quality assessment scored by GPT-4, evaluating relevance, fluency, informativeness, and logical coherence.
\end{itemize}


To reduce evaluation cost during the validation phase, all models are assigned a default G-Score, ensuring consistency for comparing different fine-tuning configurations.

\subsection{Validation Results}

We conduct extensive experiments on the Qwen2.5 series models, using both LoRA-based parameter-efficient fine-tuning and full fine-tuning for smaller models. Table \ref{tab:val_result} summarizes the results on the validation set.

\setlength{\tabcolsep}{4pt}
\begin{table}[htbp]
\centering
\caption{Performance on the validation set across different fine-tuning strategies}
\label{tab:val_result}
\resizebox{0.95\textwidth}{!}{
\begin{tabular}{l|l|ccccccc|c}
\hline
\textbf{Base Model} & \textbf{Method} & \textbf{ME} & \textbf{B-4} & \textbf{R-L} & \textbf{Ext} & \textbf{D-2} & \textbf{D-3} & \textbf{G-score} & \textbf{Total} \\
\hline
Qwen2.5-72B & w/o & 19.76 & 0.93 & 9.13 & 94.20 & 17.56 & 37.29 & 100 & 30.87 \\
Qwen2.5-72B & LoRA(8,16) & 27.38 & 7.60 & 32.58 & 96.16 & 26.49 & 44.97 & 100 & 40.28 \\
Qwen2.5-72B & LoRA(16,32) & \textbf{28.04} & \textbf{7.94} & \textbf{32.71} & 96.22 & 26.53 & 45.07 & 100 & 40.52 \\
Qwen2.5-72B & LoRA(32,64) & 26.37 & 6.29 & 29.82 & \textbf{96.11} & \textbf{32.85} & \textbf{57.82} & 100 & \textbf{41.17} \\
Qwen2.5-7B & Full & 27.87 & 6.85 & 30.63 & 96.22 & 22.21 & 39.34 & 100 & 38.85 \\
\hline
\end{tabular}
}
\end{table}

As shown in Table~\ref{tab:val_result}, all fine-tuned models outperform the base model significantly. Among them, the configuration with LoRA (rank=32, alpha=64) achieves the highest total score of 41.17.

\subsection{Test Set Submissions}

Based on validation results, we submitted two results to the test set: one trained only on the original training set and the other trained on the combined training and validation sets. Their results are shown in Table~\ref{tab:test_result}.

\begin{table}[htbp]
\centering
\caption{Performance on the test set.}
\label{tab:test_result}
\resizebox{0.95\textwidth}{!}{
\begin{tabular}{l|ccccccc|c}
\hline
\textbf{Submission ID} &  \textbf{ME} & \textbf{B-4} & \textbf{R-L} & \textbf{Ext} & \textbf{D-2} & \textbf{D-3} & \textbf{G-score} & \textbf{Total} \\
\hline
submit\_0418 &  27.06 & \textbf{7.43} & \textbf{32.06} & 96.16 & 26.22 & 44.60 & 85.17 & 38.53 \\
submit\_0420 &  \textbf{27.21} & 6.25 & 29.89 & \textbf{96.22} & \textbf{30.99} & \textbf{55.12} & \textbf{87.20} & \textbf{39.62} \\
\hline
\end{tabular}
}
\end{table}

The results show that submit\_0420, which includes additional validation data during training, achieves a higher total score of 39.62, especially improving the diversity and informativeness of generated responses.

\section{Conclusion}
This paper presents our approach to the NLPCC 2025 Task 8 on Emotional Support Conversation. By combining prompt engineering with both LoRA-based and full-parameter fine-tuning on Qwen2.5 models, we significantly improved the models' ability to generate empathetic and context-aware responses. Our best submission ranked second in the official evaluation, demonstrating the effectiveness of large language models in the ESC domain. Future work will focus on enhancing emotional understanding, personalization, and response safety to further improve user experience in real-world applications.

\bibliographystyle{splncs04}
\bibliography{nlpcc2025}




\end{document}